\documentclass[letterpaper, 10 pt, conference]{ieeeconf}  

\IEEEoverridecommandlockouts                              

\usepackage{booktabs} 
\usepackage{siunitx}  
\usepackage{hyperref}
\usepackage{cleveref}

\usepackage{graphicx}  
\usepackage{booktabs}  
\usepackage{multirow}  
\usepackage{makecell}  
\usepackage{amssymb}   
\usepackage{pifont}
\usepackage{subcaption}
\usepackage{censor}

\newcommand{\cmark}{\ding{51}}
\newcommand{\xmark}{\ding{55}}

\newcommand{\method}{\text{UniScale}~}

\makeatletter
\renewcommand{\subsection}{\@startsection{subsection}{2}{\z@}{1.5ex plus 1.5ex minus 0.5ex}%
{0.7ex plus .5ex minus 0ex}{\normalfont\normalsize\bfseries}}

\renewcommand{\paragraph}{\@startsection{paragraph}{4}{2\parindent}{0ex plus 0.1ex minus 0.1ex}%
{0ex}{\normalfont\normalsize\bfseries}}

\long\def\@makecaption#1#2{%
\ifx\@captype\@IEEEtablestring%

\setbox\@tempboxa\hbox{\footnotesize \textbf{#1.}~~ #2}%
\ifdim \wd\@tempboxa >\hsize%
\setbox\@tempboxa\hbox{\footnotesize \textbf{#1.}~~ }%
\parbox[t]{\hsize}{\footnotesize \noindent\unhbox\@tempboxa#2}%
\else%
\hbox to\hsize{\footnotesize\hfil\box\@tempboxa\hfil}%
\fi\par
\@IEEEtablecaptionsepspace%
\else
\@IEEEfigurecaptionsepspace%
\setbox\@tempboxa\hbox{\footnotesize \textbf{#1.}~~ #2}%
\ifdim \wd\@tempboxa >\hsize%
\setbox\@tempboxa\hbox{\footnotesize \textbf{#1.}~~ }%
\parbox[t]{\hsize}{\footnotesize \noindent\unhbox\@tempboxa#2}%
\else%
\ifcenterfigcaptions \hbox to\hsize{\footnotesize\hfil\box\@tempboxa\hfil}%
\else \hbox to\hsize{\footnotesize\box\@tempboxa\hfil}%
\fi\fi\fi}
\makeatother

\title{\LARGE \bf
UniScale: Unified Scale-Aware 3D Reconstruction for Multi-View Understanding via Prior Injection for Robotic Perception
}

\author{Mohammad Mahdavian$^{1}$*, Gordon Tan$^{1,2}$*$^{\dagger}$, Binbin Xu$^{1}$, Yuan Ren$^{1}$, Dongfeng Bai$^{1}$, Bingbing Liu$^{1}$ 
\thanks{ * Equal Contribution}
\thanks{{$\dagger$} Work done during an internship at Huawei Noah’s Ark Lab}
\thanks{$^{1}$Huawei Noah’s Ark Lab.
         \tt\small \{mohammad.mahdavian1, } 
\thanks{\tt\small binbin.xu, yuan.ren3, baidongfeng,    liu.bingbing\}@huawei.com}
\thanks{$^{2}$University of Toronto,
        \tt\small gordon.tan@mail.utoronto.ca}%
}

\begin{document}

\maketitle
\thispagestyle{empty}
\pagestyle{empty}

\begin{abstract}

We present UniScale, a unified, scale-aware multi-view 3D reconstruction framework for robotic applications that flexibly integrates geometric priors through a modular, semantically informed design. In vision-based robotic navigation, the accurate extraction of environmental structure from raw image sequences is critical for downstream tasks. UniScale addresses this challenge with a single feed-forward network that jointly estimates camera intrinsics and extrinsics, scale-invariant depth and point maps, and the metric scale of a scene from multi-view images, while optionally incorporating auxiliary geometric priors when available.
By combining global contextual reasoning with camera-aware feature representations, UniScale is able to recover the metric-scale of the scene. In robotic settings where camera intrinsics are known, they can be easily incorporated to improve performance, with additional gains obtained when camera poses are also available. This co-design enables robust, metric-aware 3D reconstruction within a single unified model. Importantly, UniScale does not require training from scratch, and leverages world priors exhibited in pre-existing models without geometric encoding strategies, making it particularly suitable for resource-constrained robotic teams. We evaluate UniScale on multiple benchmarks, demonstrating strong generalization and consistent performance across diverse environments. We will release our implementation upon acceptance.

\end{abstract}

\section{INTRODUCTION}

Accurate 3D scene reconstruction plays a central role in robotic perception, enabling core tasks such as navigation, mapping, and interaction. While recent learning-based multi-view methods have demonstrated impressive performance using raw images, their deployment in the wild is often hindered by scale ambiguity, rigid architectures, and high computational costs. To operate effectively under diverse sensing conditions, robotic systems benefit significantly from reconstruction models that are adaptable and metric-aware.

Historically, image-based 3D reconstruction has progressed by decomposing the task into individual components, including depth estimation~\cite{eigen2014depth}, camera calibration~\cite{zhou2017unsupervised}, and point cloud generation~\cite{fan2017point}. Classical approaches, including Structure from Motion (SfM)~\cite{pan2024global} and Multi-View Stereo (MVS)~\cite{yao2018mvsnet}, typically rely on multi-stage pipelines with carefully engineered optimization to achieve these results.

Recent learning-based approaches adopt unified feed-forward architectures that jointly predict depth, point clouds, and camera poses from images. Models such as VGGT~\cite{wang2025vggt}, MV-DUSt3R~\cite{tang2025mv}, MUSt3R~\cite{cabon2025must3r}, and MASt3R~\cite{leroy2024grounding} demonstrate strong generalization and efficiency, making them promising for robotics. However, many still struggle with reliable metric-scale recovery and flexible incorporation of geometric priors, both critical for real-world deployment. Pow3R~\cite{jang2025pow3r} integrates geometric priors (e.g., intrinsics and poses) into a unified representation, benefiting tasks such as robotic scene understanding and exploration~\cite{alama2025rayfronts}, yet it produces scale- or affine-invariant outputs rather than fully metric reconstructions.

\begin{figure}[t]
\vspace{-2mm}
  \centering
  \includegraphics[width=\linewidth]{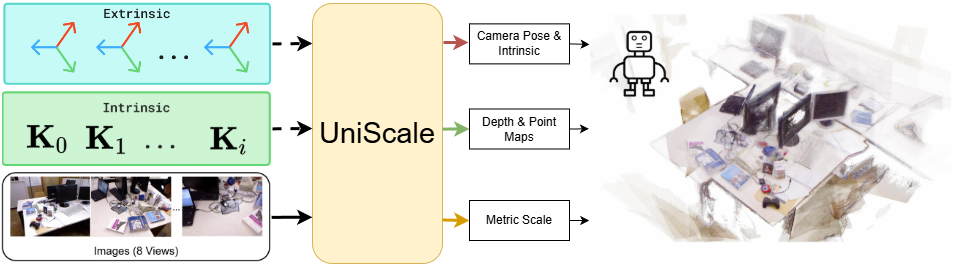}
  \caption{UniScale Overview. Upon receiving a set of images with optional camera intrinsic and extrinsic information, UniScale generates depth and point maps, metric-scale and auxilliary camera information, all of which may be used in 3D reconstruction for downstream robotic tasks. }
  \label{fig:cool}
  \vspace{-4mm}
\end{figure}

Metric depth estimation is essential for many robotic applications, yet recovering absolute scale from monocular images is inherently ill-posed. Recent foundation models mitigate this challenge through large-scale training, enabling metric depth and point cloud prediction from single views, as shown by Metric3D~\cite{yin2023metric3d}, Metric3D-V2~\cite{hu2024metric3d}, and MoGe-2~\cite{wang2025moge2}. MapAnything~\cite{keetha2025mapanything} further unifies depth estimation, camera inference, and scale recovery while supporting geometric priors. However, forcefully embedding all prior information into the same image features may significantly limit the exploitation of structured geometry, and training such models from scratch remains computationally expensive for many robotic settings.

We build UniScale on top of VGGT~\cite{wang2025vggt}, as shown in Fig.\ref{fig:cool}, extending it to support a broader range of multi-view 3D reconstruction tasks, including metric-scale prediction and geometric prior injection. We focus on incorporating camera intrinsics and poses, which are available in most robotic systems and can be reliably integrated during reconstruction. 

Our model addresses key limitations of prior works. First, we introduce a dedicated metric-scale head to estimate real-world scene scale, overcoming the scale invariance of models such as VGGT~\cite{wang2025vggt}. Second, unlike the uniform prior injection in MapAnything~\cite{keetha2025mapanything}, UniScale adopts semantic-aware prior injection, distributing priors across embedded tokens according to their semantic roles. This dual-component design improves upon VGGT and outperforms MapAnything on several benchmarks. Moreover, UniScale’s modular architecture enables seamless integration into other unified 3D perception frameworks, providing metric-scale estimation and flexible prior injection for broader robotic applicability.

In summary, our main contributions are as follows:
\begin{itemize}
    \item We propose a unified framework for multi-view metric 3D reconstruction that supports the injection of camera priors for robotic applications.
    \item We introduce a modular metric-scale head that recovers real-world scale by fine-tuning globally learned features, leveraging image, camera, and aggregated patch tokens from VGGT.
    \item We design a semantic-aware prior injection mechanism that adapts the injection process according to the role of each embedding.
    \item Our modular design enables seamless integration into diverse robotic 3D reconstruction frameworks.
    \item We achieve competitive or superior performance compared to prior unified and single-task feed-forward approaches across multiple benchmarks.
\end{itemize}

\vspace{-1mm}

\section{Related Works}

\subsection{Multi-View 3D Reconstruction}

Recent 3D vision methods emphasize feed-forward reconstruction from image pairs or sequences. DUSt3R~\cite{wang2024dust3r} predicts dense 3D point maps without explicit matching or calibrated intrinsics, while MASt3R~\cite{leroy2024grounding} and MUSt3R~\cite{cabon2025must3r} extend it with dense correspondence and scalable multi-view modeling. VGGT~\cite{wang2025vggt} further adopts a transformer-based framework for joint visual–geometric reasoning in non-metric reconstruction. Building on this unified paradigm, our method introduces a metric-scale head and a contextual prior-injection mechanism to enhance reconstruction fidelity.

\subsection{3D Reconstruction with Prior Injection}

Geometric priors such as intrinsics and poses enhance 3D reconstruction robustness. Existing methods either inject priors as tokens (e.g., Pow3R~\cite{jang2025pow3r}, G-CUT3R~\cite{khafizov2025g}) or embed them directly into attention operations (e.g., CAPE~\cite{Kong_2024_CVPR}, GTA~\cite{Miyato2024GTA}, PRoPE~\cite{li2025cameras}). We adopt a token-based injection for modularity, and introduce a semantic-aware strategy that routes information from each prior to the most relevant tokens(e.g., pose to camera tokens, intrinsics to patch tokens).

The parameterization of priors is critical for stable convergence. While quaternions are commonly used for camera extrinsics~\cite{wang2025vggt, keetha2025mapanything}, their discontinuities can hinder optimization; we instead adopt a continuous 6D pose representation~\cite{zhou2019continuity}. For intrinsics, we use a ray-based encoding that captures camera geometry explicitly. Dedicated encoders project these priors into the token space, favoring information richness over numerical simplicity.

\subsection{3D Reconstruction for Robotics}
Recent works leverage 3D reconstruction for robotic perception and task execution. OpenNavMap~\cite{jiao2026opennavmap} employs MASt3R for structure-free topometric mapping, while Chi et al.~\cite{chi2026dynamic} integrate 3D modeling with path optimization for autonomous humanoid welding. Task-specific systems such as SurfSLAM~\cite{bagoren2026surfslam} and AgriGaussian~\cite{an2026agrigaussian} enable metric reconstruction for underwater navigation and agricultural phenotyping, respectively. Additionally, recent approaches such as StreamVGGT~\cite{zhuo2025streaming} and InfiniteVGGT~\cite{yuan2026infinitevggt} extend 3D reconstruction to streaming settings, 
their potential for robotic applications. Nevertheless, many existing methods struggle to achieve reliable metric-scale estimation or to flexibly incorporate geometric priors. In contrast, our approach overcomes these limitations. Operating in real time within a sliding-window framework, it produces metric-scale reconstructions and naturally integrates geometric priors, making it well suited for downstream robotic tasks.


\begin{figure*}[t]
  \centering
  \begin{subfigure}[t]{0.64\textwidth}
    \centering
    \includegraphics[width=\linewidth]{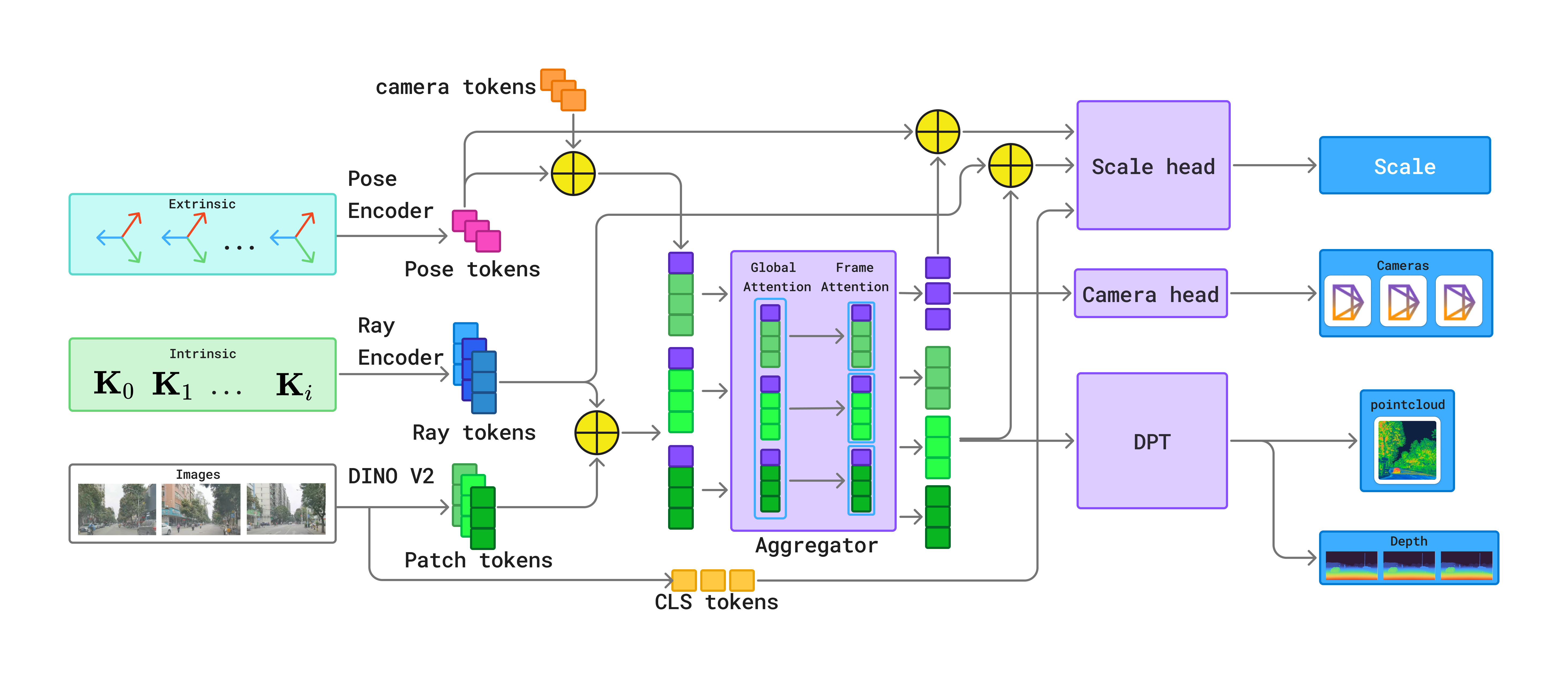}
    \vspace{-9mm}
    \caption{}\label{fig:uniscale_structure}
  \end{subfigure}
  \hfill
  \begin{subfigure}[t]{0.34\textwidth}
    \centering
    \raisebox{11mm}{\includegraphics[width=\linewidth]{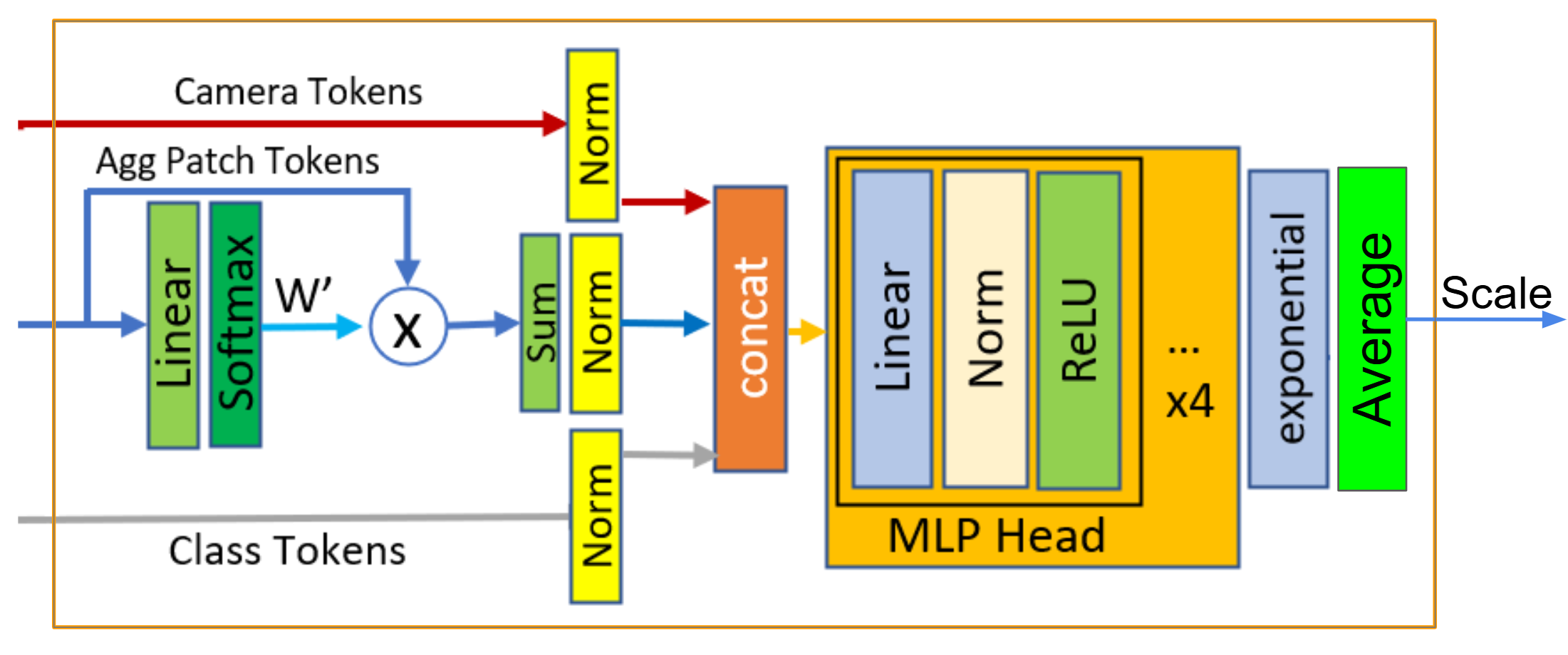}} 
    \vspace{-9mm}
    \caption{}\label{fig:scale_head}
  \end{subfigure}
\vspace{-3mm}
  \caption{(a) Overview of the UniScale architecture and (b) Architecture of the Scale Head. The model combines global contextual information from class tokens, camera intrinsics and extrinsics encoded in camera tokens, and image features from aggregated patch tokens to predict the scene-level scale value.}
  \label{fig:uniscale_pipeline}
  \vspace{-5mm}
\end{figure*}

\subsection{Monocular Depth or Point Map Estimation}

3D reconstruction methods can be broadly divided into metric and non-metric approaches. Early monocular models typically produced scale- or affine-invariant depth due to inherent ambiguities, as exemplified by MiDaS~\cite{ranftl2020towards}, MoGe~\cite{wang2025moge}, and Marigold~\cite{ke2024repurposing}. Prior work addressed this by incorporating additional cues such as sparse SfM depth~\cite{ma2018sparse}, ground-plane constraints~\cite{wagstaff2021self}, and multi-view geometry~\cite{wei2023surrounddepth}.

Recent foundation models enable metric-scale estimation from large-scale training. Metric3D~\cite{yin2023metric3d}, MoGe-2~\cite{wang2025moge2}, and MapAnything~\cite{keetha2025mapanything} recover scale through specialized normalization or scale heads. In contrast, our method introduces a dedicated scale head that refines globally learned features and integrates with scale-invariant predictors, enabling metric reconstruction for downstream robotic tasks.


\section{UniScale}

We introduce UniScale, a unified framework for metric 3D reconstruction and multi-view understanding with the capability of incorporating priors when available. We begin by defining the problem in~\cref{sec:def} and describing the main model architecture in~\cref{sec:main_model}. Next, we present our prior injection mechanism in~\cref{sec:prior} and the metric-scale prediction paradigm in~\cref{sec:metric}. Finally, we detail the UniScale training procedure in~\cref{sec:train}.

\subsection{Problem Definition}\label{sec:def}

Our proposed approach is an end-to-end model for 3D scene reconstruction suitable for robotic applications that takes as input a sequence of consecutive or cluttered RGB images, $(I_i)_{i=1}^N$, $I_i\in \mathbb{R}^{3\times H\times W }$, and our transformer-based model jointly predicts scale-invariant depth maps, $(D_i)_{i=1}^N$, $D_i\in \mathbb{R}^{H\times W }$, 3D pointclouds, $(PC_i)_{i=1}^N$, $PC_i\in \mathbb{R}^{3 \times H \times W }$, and the corresponding camera intrinsics and scale-invariant extrinsics, $(g_i)_{i=1}^N$, $g_i \in \mathbb{R}^9$. In addition, a dedicated scale head estimates the absolute metric-scale of the scene, $S \in \mathbb{R}$. A key advantage of our method is its ability to incorporate prior information, such as camera intrinsics, $(K_i)_{i=1}^N$, $K_i \in \mathbb{R}^{3\times3}$, or poses, $(P_i)_{i=1}^N$, $P_i \in \mathbb{R}^{4\times4}$ obtained from SfM or related techniques, into the UniScale model, thereby enhancing reconstruction accuracy and overall performance. The general equation for this process can be defined as~\cref{eq:main}.

\vspace{-3mm}
 
\begin{equation}\label{eq:main}
 f\!\left((I_i)_{1}^N,\!\ \mathrm{Opt}\!\left((K_i)_{1}^N,\!\ (P_i)_{1}^N\right)\right)
  = (g_i, D_i, PC_i, S)_{1}^N 
\end{equation}

The camera information, $g_i=[q_i,t_i,f_i]$, are parametrized based on~\cite{wang2024vggsfm} and combine rotation quaternion, $q \in \mathbb{R}^4$, translation, $t \in \mathbb{R}^3$, and field-of-view (fov) of the camera, $f \in \mathbb{R}^2$. 

\subsection{Model Structure}\label{sec:main_model}

Fig.~\ref{fig:uniscale_structure} illustrates the UniScale architecture, which combines backbone encoders, a global-frame aggregator, and specialized heads for depth, camera, and scale estimation. Following recent 3D reconstruction advances~\cite{keetha2025mapanything,wang2025vggt,wang2024dust3r}, we adopt large-scale transformers~\cite{vaswani2017attention} as flexible models with minimal geometric bias, enabling geometric reasoning to be learned from large, diverse datasets.

\paragraph{Image Feature Backbone}
Our transformer-based model, $f$, uses dedicated encoders to extract features from images and camera priors. 
Each image $i \in [1,\dots,N]$ is patchified using DINOv2~\cite{oquab2023dinov2} into $N_{pt}$ patch tokens $t_i^{pt} \in \mathbb{R}^{N_{pt} \times C}$ and a class token $t_i^{L} \in \mathbb{R}^{1 \times C}$, capturing local and global features.
Following VGGT~\cite{wang2025vggt}, we concatenate the patch tokens with a learnable camera token $t_i^{g} \in \mathbb{R}^{1 \times C}$ and four register tokens $t_i^{R} \in \mathbb{R}^{4 \times C}$ before aggregation. The camera token enables intrinsic and extrinsic estimation, while register tokens improve stability~\cite{darcet2023vision}. For the first frame, distinct camera and register tokens anchor all 3D predictions to its coordinate system.

\paragraph{Aggregator}

The aggregator comprises a global attention module for cross-frame interactions and a frame-level attention module for intra-frame dependencies~\cite{wang2025vggt}. This design encodes both local details and global geometric context for multi-view reasoning. Its outputs are the processed camera tokens $(\hat{t}_i^{g})_{i=1}^{N}$, with $\hat{t}^{g} \in \mathbb{R}^{1 \times C}$, and the aggregated patch tokens $(\hat{t}_i^{pt})_{i=1}^{N}$, with $\hat{t}^{pt} \in \mathbb{R}^{N_{pt} \times C}$.

\paragraph{Prediction Heads}

Our model jointly predicts camera intrinsics and extrinsics, scale-invariant depth and point maps, and metric-scale (see~\cref{sec:metric}). 
The camera head designates the global coordinate frame using the first frame’s tokens, and predicts remaining parameters from $(\hat{t}_i^{g})_{i=1}^{N}$ via self-attention and linear projection, yielding $(g_i)_{i=1}^{N}$.
For dense prediction, aggregated patch tokens are processed by a DPT head~\cite{ranftl2021vision} to produce normalized depth maps $D_i$ and point maps $PC_i$, with aleatoric uncertainty~\cite{kendall2016modelling}. To obtain metric translation, depth maps, and point maps, we simply multiply them by the predicted scale factor.

\subsection{Prior Injection} \label{sec:prior}
\label{subsec:prior}
Our model supports conditioning on camera intrinsics and extrinsics through two dedicated encoders: a pose encoder and a raymap encoder. As illustrated in Fig.\ref{fig:uniscale_pipeline}, we inject pose embeddings into the camera tokens and scale head, and ray embeddings into the patch tokens. This semantic-aware routing directs relevant priors to appropriate heads while minimizing noise.

\paragraph{Pose Encoder} We employ a simple MLP as our pose encoder. The input is a pose tensor to each frame, $G_i \in \mathbb{R}^9$ consisting of the concatenated rotation and translation of the frame, such that: 

\vspace{-4mm}
\begin{equation} \label{pose_concat}
    G_i =  (R_i, T_i)
\end{equation} 

\noindent where $R_i \in \mathbb{R}^6$ and $T_i \in \mathbb{R}^3$ represent the frames' rotation and translation.  For faster and more stable convergence during training, we parameterize our rotation matrices using a 6D representation following \cite{zhou2019continuity} such that $R_i \in \mathbb{R}^6$. This representation is by nature continuous, and, due to their orthogonal property, they are also bijective with respect to the rotation space. We observed that this aids in faster convergence compared with training with quaternions which are discontinuous. Our output pose encoding is of the shape of the camera tokens, $t^g_i$ and then directly perform element-wise addition to obtain our final input camera tokens.


\paragraph{Intrinsics Encoder} 

Following Pow3R~\cite{jang2025pow3r}, we encode camera intrinsics as origin-free ray images, $Rays \in \mathbb{R}^{H \times W \times 3}$. We found that including origin information, as in Plücker rays~\cite{plucker1865new} or standard raymaps, introduces unnecessary noise during training and inference. The resulting embeddings match the patch token dimension $t^{pt}$ and are added element-wise.

\vspace{-1mm}
\subsection{Metric-Scale Prediction} \label{sec:metric}

In our metric-scale head illustrated in Fig.~\ref{fig:scale_head}, the model predicts a scene-level scale value $S$ conditioned on the observed visual content and, when available, pose and intrinsic priors. We combine multiple processed tokens that provide geometric cues. The class tokens $t^L$ produced by DINOv2 capture rich high-level contextual information regarding the scene, while the camera tokens $\hat{t}^{g}$ from the aggregator encode camera intrinsics and extrinsics, facilitating accurate scale estimation. Additionally, the aggregated patch tokens $\hat{t}^{pt}$ capture inter- and intra-frame relationships, further enhancing the robustness of metric-scale prediction. When available, we also incorporate pose and ray embeddings as priors, enabling the scale head to estimate the scene scale with higher accuracy. The Scale Head can be defined as~\cref{scaleHead_eq}.

\vspace{-2mm}
\begin{equation} \label{scaleHead_eq}
 S = ScaleHead\!\left(\hat{t}^{g}, \hat{t}^{pt}, t^L, \mathrm{Opt}(K,\!\ P)\right)
\end{equation}

\vspace{-1mm}

We first downsample our $N_{pt}$ patch tokens before merging them with other input tokens. To achieve this, we introduce a pseudo-attention module for adaptive downsampling. First, a linear layer, $l_1$, projects the $C$-dimensional embeddings of each patch to a single scalar, generating weights $W \in \mathbb{R}^{N_{pt} \times 1}$. A softmax function is then applied to $W$ for normalization. These weights are then used to compute a weighted sum over the patch dimension of $\hat{t}^{pt}$, producing a downsampled patch token $\hat{W} \in \mathbb{R}^{C}$ as in~\cref{agg_ds_eq}.

\vspace{-2mm}
\begin{equation} \label{agg_ds_eq}
\hat{W} = \left(\sum_{p=1}^{N_{pt}} 
    \mathrm{Softmax}\!\bigl(l_1(\hat{t}^{pt})\bigr)_p\right)\, \times \hat{t}^{pt}
\end{equation}
\vspace{-3mm}

Next, we normalize the class, camera, and downsampled patch tokens and concatenate them along the feature dimension to form $T$, as shown in~\cref{T_eq}. This normalization step stabilizes the performance of the scale head and ensures a balanced contribution from all feature sources. The combined representation $T$ is then passed through a MLP. Finally, an exponential activation is applied to the output and averaged over the frames to produce the predicted scale value $S$ of the scene, as illustrated in~\cref{S_eq}.

\vspace{-4mm}

\begin{equation}\label{T_eq}
  T = \mathrm{Concat}\!\left(
        \mathrm{Norm}(\hat{t}^{g}),
        \mathrm{Norm}(\hat{t}^{pt}),
        \mathrm{Norm}(t^{L})
      \right)
\end{equation}

\vspace{-3mm}

\begin{equation}\label{S_eq}
  S = \frac{1}{N}\sum_{i=1}^{N} \exp\bigl(\mathrm{MLP}(T_i)\bigr)
\end{equation}
\vspace{-2mm}

To capture metric-scale information from the injected priors, we also integrate them into the metric-scale head. We incorporate the pose embeddings into the camera tokens before normalization and the ray embeddings into the aggregated patch tokens prior to the pseudo-attention layer. These design choices are motivated by both contextual relevance and structural similarity: pose embeddings align naturally with the camera tokens in terms of semantics and feature dimensionality, whereas ray embeddings correspond more closely to the spatial structure and feature shape of the aggregated patch tokens.

\subsection{Training Unified Metric 3D Reconstruction}\label{sec:train}

\paragraph{Training Loss} We train the UniScale model end-to-end in a multi-task setting, employing multiple loss functions, each responsible for optimizing a specific component of the model. The total loss is formulated as:

\vspace{-2mm}

\begin{equation}
\mathcal{L} = \mathcal{L}_{\text{camera}} + \mathcal{L}_{\text{depth}} + \mathcal{L}_{\text{pmap}} + \mathcal{L}_{\text{scale}} 
\end{equation}

Following VGGT~\cite{wang2025vggt}, we adopt similar formulations for camera, depth, and point map (pmap) losses. 
For the camera loss, $\mathcal{L}_{\text{camera}}$, we employ a Huber loss between the predicted camera parameters and their ground-truth values. For the depth loss, $\mathcal{L}_{\text{depth}}$, we adopt the aleatoric uncertainty formulation~\cite{kendall2017uncertainties}, which weights the discrepancy between the predicted and ground-truth depth by the model’s predicted uncertainty map. We additionally include a gradient-based term to preserve edge sharpness in both the depth and point maps. The $\mathcal{L}_{\text{pmap}}$ is calculated similarly to that of $\mathcal{L}_{depth}$, but using the uncertainty of the point map.
For scale loss, $\mathcal{L}_{scale}$, we calculate the $\ell_2$ norm on the logarithmic difference between the ground-truth $\hat{S}$ and predicted $S$ scale values, resulting in $ \mathcal{L}_{\text{scale}} = \left\| \log(\hat{S}) - \log(S) \right\|_2 $. 
Parameterizing scale in the logarithmic space allows the model to handle larger variances in magnitude, across both indoor and outdoor scenes. It is important to mention that we only train our scale head on metric datasets and mask $ \mathcal{L}_{\text{scale}} = 0$ for data samples from our non-metric datasets, CO3Dv2~\cite{reizenstein2021common} and MegaDepth~\cite{li2018megadepth}.

\paragraph{Training with Prior Injection} 

To train a single model robust to varying input configurations, we adopt a probabilistic prior injection strategy similar to MapAnything~\cite{keetha2025mapanything}. During training, different combinations of geometric priors are randomly provided, encouraging robustness to missing inputs.

Specifically, priors are injected with probability $0.5$, and each prior type (pose and intrinsics) is independently included with probability $0.9$. For datasets with metric ground truth, scale supervision is provided with probability $0.95$.

\paragraph{Implementation Details} 

We initialize the image encoder, alternating attention module, and camera, depth, and point heads with pre-trained DINOv2 and VGGT weights. UniScale is trained for 38K iterations using AdamW~\cite{kingma2014adam} with a linear warmup from $1\times10^{-8}$. The scale head and pose and ray-map encoders use a peak learning rate of $5\times10^{-5}$, while other modules use $1\times10^{-6}$ to enable stable fine-tuning.

Each batch contains 2 to 24 randomly sampled images, yielding an effective batch size of 8 to 96. Images are resized to a maximum of 518 pixels with randomized aspect ratios in $[0.33,1.0]$, and augmented with color jitter and grayscale conversion. We employ gradient clipping and aggressive checkpointing with \texttt{bfloat16} precision to enhance training stability and efficiency.

\paragraph{Datasets} 
We train UniScale on 10 indoor and outdoor datasets from both synthetic and real-world sources: Argoverse2~\cite{wilson2023argoverse}, Aria Synthetic~\cite{avetisyan2024scenescript}, Co3Dv2~\cite{reizenstein2021common}, Hypersim~\cite{roberts2021hypersim}, MegaDepth~\cite{li2018megadepth}, MVS-Synth~\cite{huang2018deepmvs}, Replica~\cite{straub2019replica}, ScanNet~\cite{dai2017scannet}, ScanNet++~\cite{dai2017scannet}, and VKitti~\cite{cabon2020virtual}.
We correct LiDAR--RGB misalignments in Argoverse2~\cite{wilson2023argoverse} using the refinement procedure of MoGe-2~\cite{wang2025moge2}. Synthetic datasets are included to improve boundary sharpness and depth quality. Following MapAnything~\cite{keetha2025mapanything}, we exclude the ScanNet++ scenes reserved for dense multi-view benchmarking from the training set. For Robust-MVD, we retain ScanNet to match VGGT’s pretraining distribution.




\paragraph{Scale Value Calculation} 
To obtain scale-invariant point clouds, depth maps, and camera extrinsics, we follow the normalization procedure of VGGT~\cite{wang2025vggt}. We first use metric depth and intrinsics to generate local 3D point clouds, which are transformed into a global cloud using the predicted extrinsics. The cloud is then normalized so that the average point distance to the origin equals one. To reduce noise, we cap the maximum depth values. The resulting normalization factor defines the supervision for training our scale head.

\vspace{0mm}
\section{Results and Benchmarking}
\vspace{-1mm}
In this section, we compare our method to the state-of-the-art (SOTA) approaches across multiple task benchmarks to show its capabilities. We perform all tasks with the same checkpoint and configurations.

\begin{table} [t]
\vspace{2mm}
  \centering
  \scriptsize
  \caption{Comparison with SOTA methods on the Robust-MVD benchmark~\cite{schroppel2022benchmark} using KITTI~\cite{geiger2013vision} and ScanNet~\cite{dai2017scannet}. We evaluate (a) multi-view metric prediction (lower is better) and (b) multi-view with alignment (lower is better for rel, higher is better for $\tau$), under different combinations of known intrinsics ($K$) and poses.} \vspace{-2mm}
\begin{tabular}{p{2.4cm}p{0.3cm}p{0.3cm}p{0.5cm}p{0.5cm}p{0.5cm}p{0.5cm}}
 \hline
  &  &   & \multicolumn{2}{c}{\textbf{KITTI}} & \multicolumn{2}{c}{\textbf{ScanNet}}\\
\textbf{Approach}  & \textbf{K} & \textbf{Poses}  & {rel $\downarrow$} & {$\tau\uparrow$} & {rel $\downarrow$} & {$\tau\uparrow$} \\
\hline 

\multicolumn{7}{l}{\textbf{a) Multi-View Metric}} \\
MAST3R~\cite{leroy2024grounding}  & \xmark & \xmark &  61.4 & 0.4 & 12.80 & 19.4 \\
MUSt3R~\cite{cabon2025must3r}  & \xmark & \xmark &  19.76 & 7.3 & 7.66 & 35.7 \\
MapAnything~\cite{keetha2025mapanything}  & \xmark &  \xmark & 5.67 & 42.7 & 32.26 & 7.1 \\
\textbf{UniScale} & \xmark & \xmark  & \textbf{5.19} & \textbf{49.6} & \textbf{5.68} & \textbf{44.5} \\

\hline
Robust MVDB~\cite{schroppel2022benchmark} & \cmark & \cmark &  7.10 & 41.9 & 7.40 & 38.4 \\
MAST3R Tri~\cite{izquierdo2025mvsanywhere} & \cmark & \cmark &  3.40 & 66.6 & 4.50 & \textbf{63.0} \\
MVSA~\cite{izquierdo2025mvsanywhere} & \cmark & \cmark &  \textbf{3.20} & \textbf{68.8} & \textbf{3.70} & 62.9 \\
MapAnything~\cite{keetha2025mapanything} & \cmark & \cmark &  4.23 & 56.4 & 17.31 & 11.9 \\
\textbf{UniScale} & \cmark & \cmark & 5.21 & 49.3 & 5.51 &  46.9\\
\hline
MapAnything~\cite{keetha2025mapanything} & \cmark & \xmark & 5.80 & 42.4 & 38.48 &  3.7 \\
\textbf{UniScale} & \cmark & \xmark & \textbf{5.16} & \textbf{49.4} & \textbf{5.32} & \textbf{46.9} \\
\hline
MapAnything~\cite{keetha2025mapanything} & \xmark & \cmark & \textbf{4.38} & \textbf{54.4} & 20.05   & 12.0 \\
\textbf{UniScale} & \xmark & \cmark & 5.03 & 51.1 & \textbf{6.04} &  \textbf{42.7}\\
\hline
\multicolumn{7}{l}{\textbf{b) Multi-View w/ Alignment}} \\
MAST3R~\cite{leroy2024grounding}  & \xmark & \xmark &  3.30 & 67.7 & 4.30 & 64.0 \\
MUSt3R~\cite{cabon2025must3r}  & \xmark & \xmark &  4.47 & 56.7 & 3.22 & 69.2 \\
MapAnything~\cite{keetha2025mapanything}  & \xmark & \xmark &  4.07 & 58.0 & 3.96 & 60.7 \\
VGGT~\cite{wang2025vggt}  & \xmark & \xmark &  4.60 & 53.0 & 2.34 & 80.6 \\
$\pi^3$~\cite{keetha2025mapanything}  & \xmark & \xmark &  \textbf{3.09} & \textbf{69.5} & 1.98 & 83.6 \\
\textbf{UniScale} & \xmark & \xmark & 3.47 & 65.9 & \textbf{1.69} & \textbf{86.8} \\
\hline
DeMoN~\cite{ummenhofer2017demon}  & \cmark & \xmark & 15.50 & 15.2 & 12.00 & 21.0 \\
DeepV2D KITTI~\cite{teed2018deepv2d}  & \cmark & \xmark &  4.47 & 56.7 & 3.22 & 69.2 \\
DeepV2D ScanNet~\cite{teed2018deepv2d}  & \cmark & \xmark &  10.00 & 36.2 & 4.40 & 54.8 \\
MapAnything~\cite{keetha2025mapanything}  & \cmark & \xmark &  3.96 & 59.5 & 3.60 & 64.6 \\
\textbf{UniScale} & \cmark & \xmark & \textbf{3.49} & \textbf{66.1} &  \textbf{1.71} & \textbf{86.3} \\
\hline
MapAnything~\cite{keetha2025mapanything} & \xmark & \cmark & 4.06 & 58.1 & 6.82 & 38.8\\
\textbf{UniScale} & \xmark & \cmark & \textbf{3.51} & \textbf{64.7} & \textbf{1.70} & \textbf{86.7} \\
\hline
MapAnything~\cite{keetha2025mapanything} & \cmark & \cmark & 3.90 & 60.8 & 4.99  &  49.9\\
\textbf{UniScale} & \cmark & \cmark & \textbf{3.50} & \textbf{65.3} & \textbf{1.71} & \textbf{86.2}\\
\hline

\end{tabular}\vspace{-2mm}
\label{tab:RMVD}
\end{table}

\subsection{Multi-View Depth Estimation}

We evaluate UniScale on the Robust-MVD benchmark~\cite{schroppel2022benchmark}, comparing against SOTA methods for multi-view metric and median-aligned depth estimation. Results are reported in Table~\ref{tab:RMVD} on KITTI~\cite{geiger2013vision} and ScanNet~\cite{dai2017scannet} under different input configurations.
UniScale achieves SOTA performance in several settings. For image-only metric prediction (a), it outperforms all prior methods on both datasets. With intrinsic priors (b), it attains SOTA results on both rel $\downarrow$ and $\tau \uparrow$. In the image-only median-aligned setting, UniScale is SOTA on ScanNet and competitive with $\pi^3$~\cite{keetha2025mapanything} on KITTI.
In other configurations, UniScale remains robust and competitive. Incorporating pose priors yields SOTA median-aligned performance and improves metric prediction on ScanNet, demonstrating effective use of pose cues. When all priors are available, UniScale achieves SOTA median-aligned results and remains competitive with specialized methods such as MVSA~\cite{izquierdo2025mvsanywhere}. Overall, these results highlight UniScale’s flexibility and effectiveness across both evaluation protocols.

\subsection{Multi-View Dense Reconstruction}

\begin{figure}[t]
  \centering
  \includegraphics[width=8.5cm]{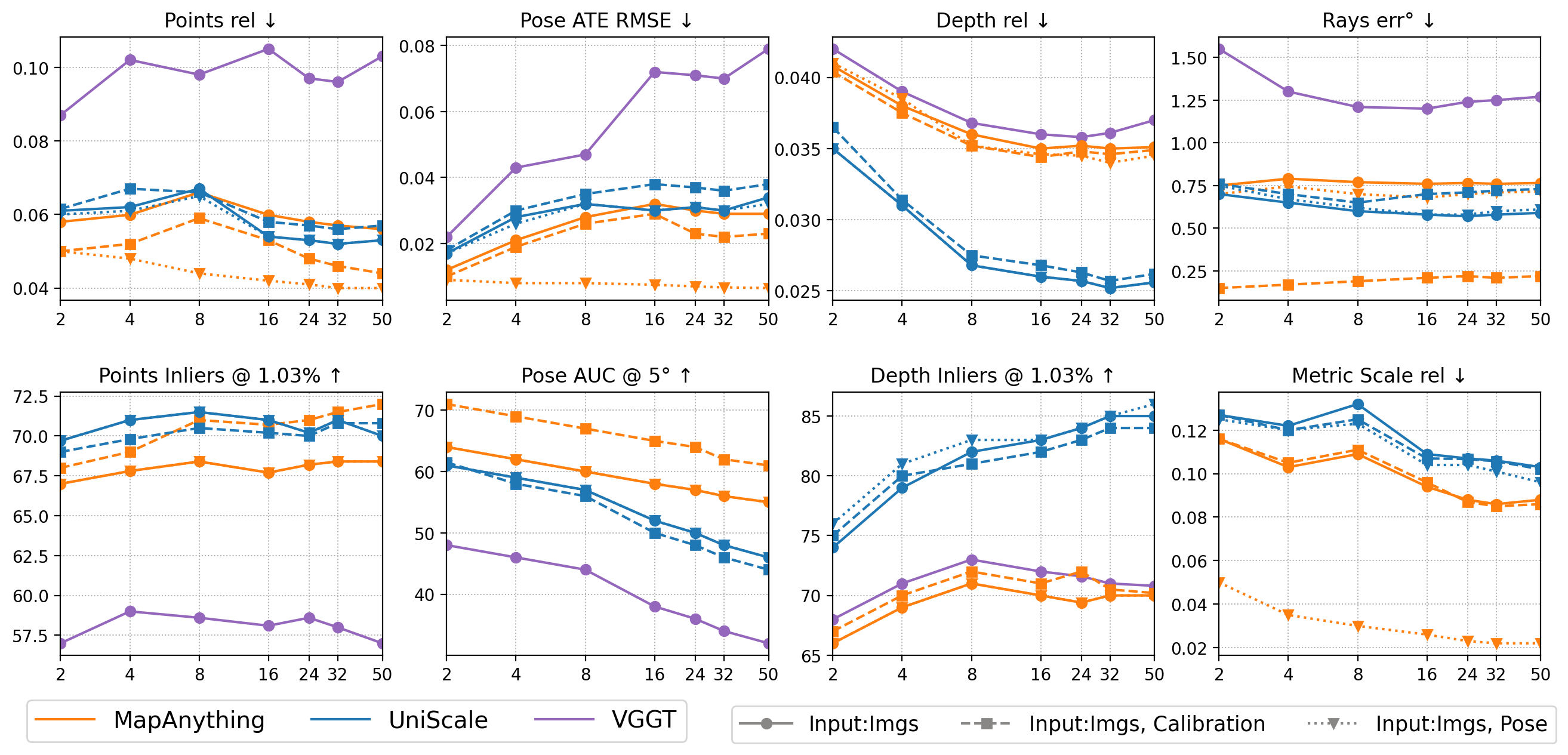}
  \vspace{-2mm}
  \caption{Comparison between UniScale and other SOTA methods on modified dense-$N$-view benchmark. UniScale demonstrates better or comparable dense multi-view reconstruction for number of input views varying from 2 to 50.}
  \label{fig:dense_n_view} \vspace{-2mm}
\end{figure}

We evaluate UniScale on the dense-$N$-view benchmark (Fig.~\ref{fig:dense_n_view}) introduced in MapAnything~\cite{keetha2025mapanything} using ETH3D~\cite{schops:etal:CVPR2017} and ScanNet++~\cite{dai2017scannet}. We report performance on point map, pose, depth, and ray direction estimation, and exclude TartanAirV2 for fairness, as it is small, designed for two-view depth, and not used in our training.
For each test, up to $N$ covisible views are randomly sampled. As shown in Fig.~\ref{fig:dense_n_view}, UniScale consistently outperforms VGGT~\cite{wang2025vggt} and surpasses MapAnything across multiple metrics, with particularly strong gains in depth estimation. UniScale achieves the lowest depth estimation error by a large margin, which is critical for robotic applications, and maintains superior ray accuracy and point inlier rates under both image-only and image+pose settings, while intrinsics slightly degrade performance. Although MapAnything performs marginally better on scale estimation in the image-only setting and benefits more from pose injection on metric scale and point map metrics, UniScale demonstrates clear advantages in depth quality and geometric consistency. Notably, as shown in Table~\ref{tab:RMVD}, UniScale’s pose injection achieves superior performance under median-aligned evaluation, while its metric depth prediction remains consistently strong on Robust-MVD. On other metrics, UniScale remains competitive. Moreover, MapAnything is trained from scratch, whereas UniScale fine-tunes a pretrained VGGT model, highlighting its practical advantages and modularity..

\begin{figure}[t]
  \centering
  \includegraphics[width=\linewidth]{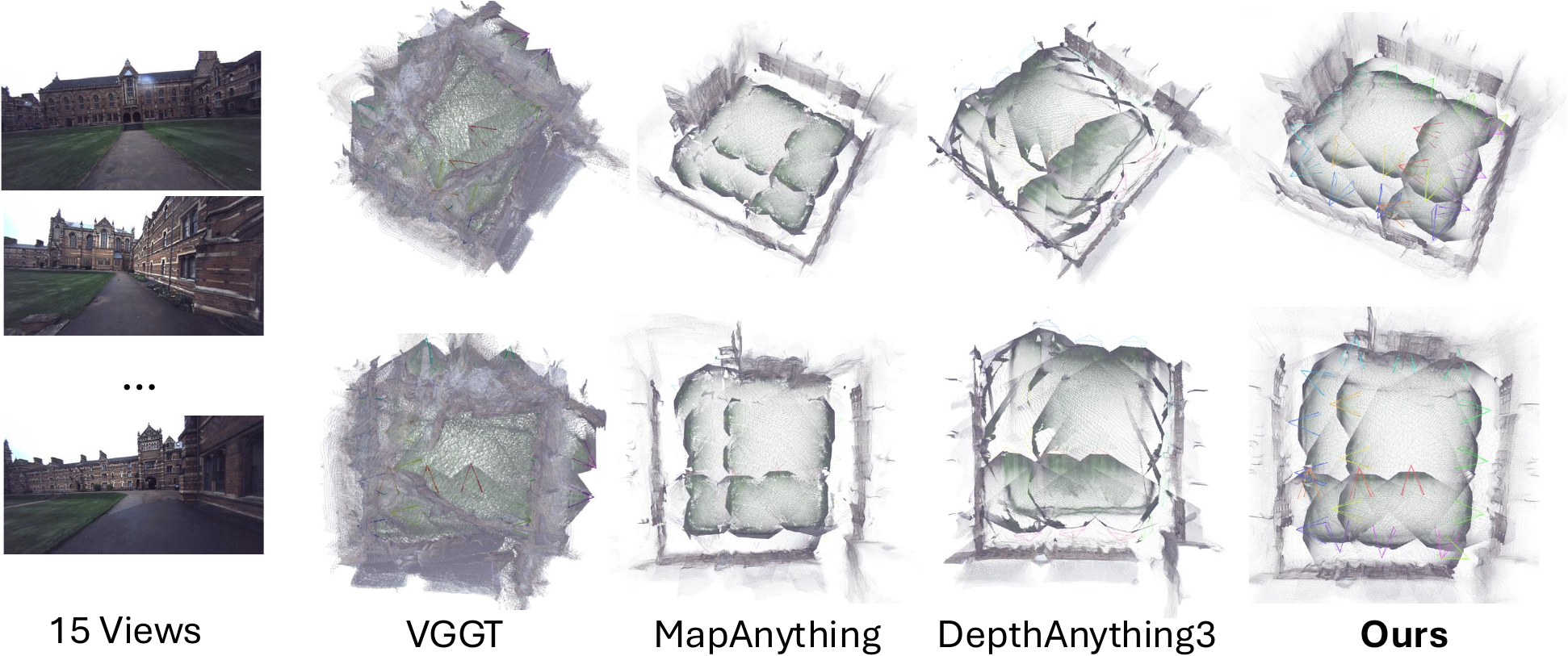}
  \caption{Qualitative Comparison - Oxford Spires Dataset~\cite{tao2025spires}}\vspace{-4mm}
  \label{fig:oxford_spires}
\end{figure}


\subsection{In-the-Wild Image 3D Reconstruction}

\begin{figure}[t]
  \centering
  \includegraphics[width=\linewidth]{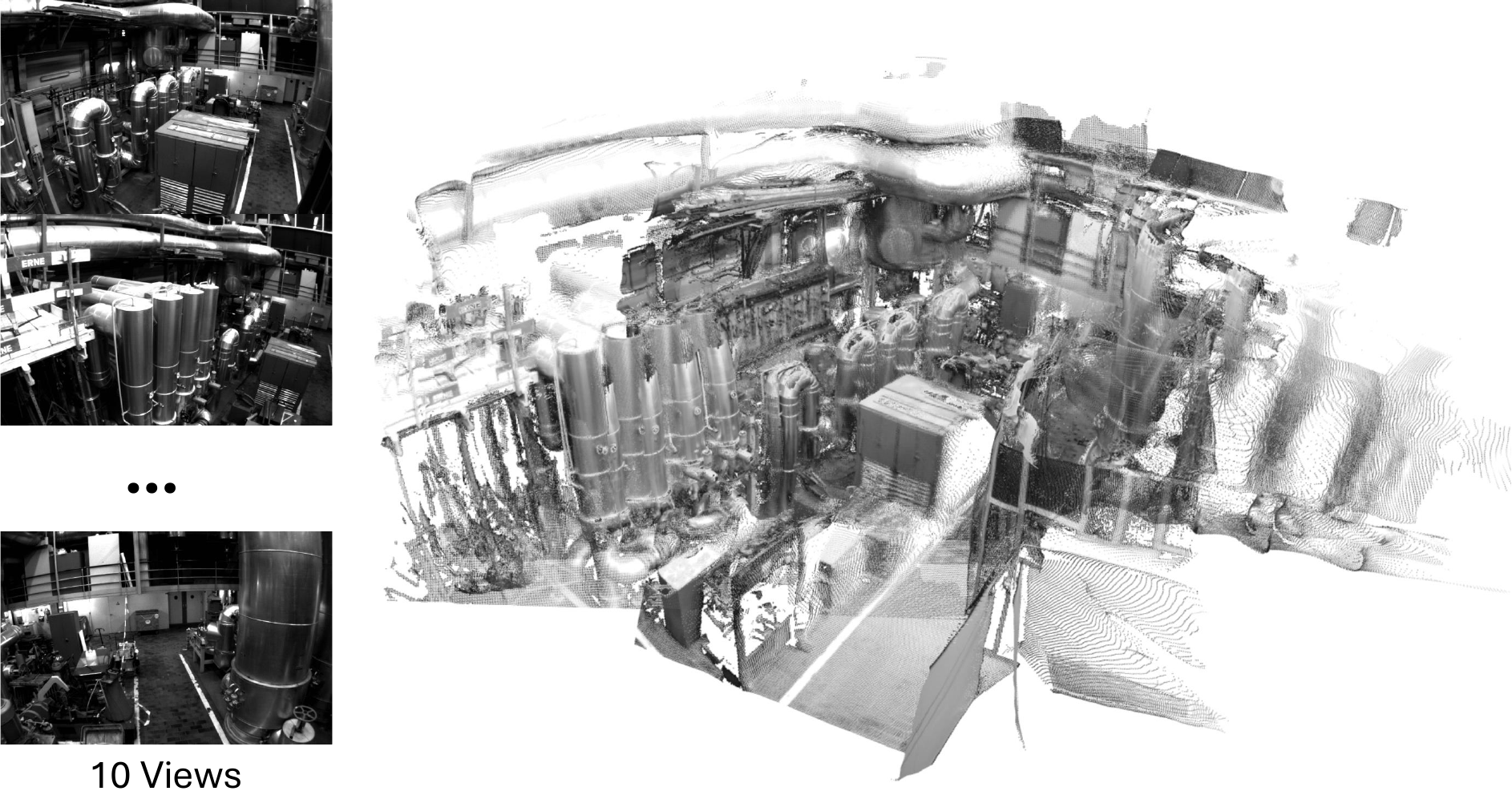}
  \caption{Qualitative Reconstruction-EuRoC MAV Dataset~\cite{Burri:etal:IJRR2016}}
   \vspace{-3mm}
  \label{fig:euroc}
\end{figure}

We evaluate the generalization of \method on unseen datasets, including EuRoC MAV~\cite{Burri:etal:IJRR2016}, TUM RGBD~\cite{Sturm:etal:IROS2012}, and Oxford Spires~\cite{tao2025spires}. Results demonstrate robust performance across diverse environments, from indoor offices~(Fig.~\ref{fig:cool}) to large-scale outdoor scenes~(Figs.~\ref{fig:oxford_spires},~\ref{fig:euroc}).

Moreover, Fig.~\ref{fig:oxford_spires} presents a qualitative comparison with SOTA methods, showing that our approach produces more complete and geometrically coherent reconstructions than VGGT~\cite{wang2025vggt}, MapAnything~\cite{keetha2025mapanything}, and DepthAnything3~\cite{lin2025depth}.

\vspace{0mm}
\section{Ablation Studies}
\label{sec:ablation}
\vspace{-1mm}
This section, presents additional ablation studies that analyze our proposed UniScale framework. We conduct a focused evaluation of the design choices underlying the metric-scale head and the prior-injection encoders. 


\vspace{-1mm}
\subsection{Scale Head Ablations}
\label{subsec:scale_head_ablation}

We study the contribution of each component in the scale head, which fuses camera tokens, aggregated patch tokens, and the global class token. We perform an ablation by removing each input in turn, with results summarized in Table~\ref{tab:ablation_scale}. All variants are retrained under the same settings and evaluated on the Robust-MVD benchmark~\cite{schroppel2022benchmark}, considering metric performance with and without prior injection.

\paragraph{Camera Tokens} First, we evaluate the removal of camera tokens, substituting them with injecting pose priors directly into the class tokens. As shown in Table~\ref{tab:ablation_scale}, omitting  camera tokens leads to a significant increase in the AbsRel error in the `Images-only' and `Images+Intrinsics' settings. While the error decreases marginally in other configurations, we prioritize performance in the `Images-only' setting as it represents the most general use case. 
Consequently, we retain the camera tokens in the scale head design.

\paragraph{Class Tokens} Removing the class tokens slightly improves performance on the indoor ScanNet dataset but leads to a significant drop on the outdoor KITTI dataset. This suggests that class tokens encode global contextual information that is essential for inferring metric-scale in large-scale outdoor scenes where local cues are ambiguous.

\paragraph{Aggregated Patch Tokens} Similarly, excluding aggregated patch tokens has a notably negative impact across multiple metrics, particularly in outdoor scenes. This mirrors the behaviors observed with class tokens, suggesting that both token types contain complementary global contextual information crucial for robust metric-scale estimation.

These findings validate the design of the scale head, confirming that all three input sources are necessary for optimal performance.

\begin{table} [t]
\vspace{2mm}
  \centering
  \scriptsize
  \caption{Ablation of the Metric Scale Head on Robust-MVD. We evaluate different scale-head designs for UniScale on KITTI~\cite{geiger2013vision} and ScanNet~\cite{dai2017scannet} to analyze the contribution of each component.} 
\begin{tabular}{p{2.5cm}p{0.2cm}p{0.3cm}p{0.5cm}p{0.5cm}p{0.5cm}p{0.5cm}}
 \hline
  &  &   & \multicolumn{2}{c}{\textbf{KITTI}} & \multicolumn{2}{c}{\textbf{ScanNet}}\\
\textbf{Approach/Setting}  & \textbf{K} & \textbf{Poses}  & {rel $\downarrow$} & {$\tau\uparrow$} & {rel $\downarrow$} & {$\tau\uparrow$} \\
\hline 

\multicolumn{7}{l}{\textbf{Multi-View Metric}} \\
\hline 
\multicolumn{7}{l}{\textbf{a) Images-only}} \\

No Camera Token & \xmark & \xmark &  5.66 & 44.8 & 5.36 & 44.8 \\
No Class Token  & \xmark & \xmark &  7.06 & 43.1 & 4.65 & 51.9 \\
No Agg Patch Token & \xmark &  \xmark & 5.19 & 51.0 & 5.83 & 44.1 \\
\textbf{UniScale} & \xmark & \xmark  & 5.19 & 49.6 & 5.68 & 44.5 \\
\hline
\multicolumn{7}{l}{\textbf{b) Images+Poses+Intrinsics}} \\

No Camera Token & \cmark & \cmark &  4.79 & 51.6 & 5.35 & 43.1 \\
No Class Token  & \cmark & \cmark &  5.82 & 47.5 & 4.82 & 49.5 \\
No Agg Patch Token & \cmark & \cmark & 5.00 & 50.6 & 6.07 & 42.2 \\
\textbf{UniScale} & \cmark & \cmark & 5.31 & 49.5 & 5.51 &  46.9\\
\hline
\multicolumn{7}{l}{\textbf{c) Images+Intrinsics}} \\

No Camera Token & \cmark & \xmark & 5.43 & 46.0 & 5.41 &  43.9 \\
No Class Token & \cmark & \xmark & 5.93 & 48.6 & 4.61 &  51.0 \\
No Agg Patch Token & \cmark & \xmark & 5.27 & 49.7 & 5.84 &  44.1 \\
\textbf{UniScale} & \cmark & \xmark & 5.16 & 49.4 & 5.32 & 46.9 \\
\hline
\multicolumn{7}{l}{\textbf{d) Images+Poses}} \\
No Camera Token & \xmark & \cmark & 4.83 & 50.6 & 5.24 &  44.4 \\
No Class Token & \xmark & \cmark & 6.23 & 46.8 & 5.00 &  48.9 \\
No Agg Patch Token & \xmark & \cmark & 4.96 & 50.8 & 6.06 &  41.9\\
\textbf{UniScale} & \xmark & \cmark & 5.03 & 51.1 & 6.04 &  42.7\\
\hline
\end{tabular}
\label{tab:ablation_scale}
\vspace{-2mm}
\end{table}

\subsection{Prior Injection Ablations}

\begin{table}[t]
  \centering
  \scriptsize
  \caption{Prior Injection Ablation on Robust-MVD~\cite{schroppel2022benchmark}. We evaluate UniScale with and without prior injection and scale head as well as variants that exclude priors from the scale head, on KITTI~\cite{geiger2013vision} and ScanNet~\cite{dai2017scannet}} 
\begin{tabular}{p{3.4cm}p{0.1cm}p{0.3cm}p{0.5cm}p{0.5cm}p{0.5cm}p{0.3cm}}
 \hline
  &  &   & \multicolumn{2}{c}{\textbf{KITTI}} & \multicolumn{2}{c}{\textbf{ScanNet}}\\
\textbf{Approach/Setting}  & \textbf{K} & \textbf{Poses}  & {rel $\downarrow$} & {$\tau\uparrow$} & {rel $\downarrow$} & {$\tau\uparrow$} \\
\hline 

\multicolumn{7}{l}{\textbf{Multi-View Metric}} \\
\hline 
\multicolumn{7}{l}{\textbf{a) Images-only}} \\
No Prior Injection & \xmark & \xmark  & 5.19 & 48.6 & 5.70 & 44.8\\
No Prior Injection into Scale Head & \xmark & \xmark  & 5.12 & 49.6 & 6.61 & 38.9\\
\textbf{UniScale} & \xmark & \xmark  & 5.19 & 49.6 & 5.68 & 44.5 \\
\hline
\multicolumn{7}{l}{\textbf{b) Images+Poses+Intrinsics}} \\

No Prior Injection into Scale Head & \cmark & \cmark  & 5.35 & 50.2 & 6.18 & 40.4\\
\textbf{UniScale} & \cmark & \cmark & 5.31 & 49.5 & 5.51 &  46.9\\
\hline
\multicolumn{7}{l}{\textbf{c) Images+Intrinsics}} \\
No Prior Injection into Scale Head & \cmark & \xmark  & 5.35 & 48.4 & 6.09 & 40.5\\
\textbf{UniScale} & \cmark & \xmark & 5.16 & 49.4 & 5.32 & 46.9 \\
\hline
\multicolumn{7}{l}{\textbf{d) Images+Poses}} \\
No Prior Injection into Scale Head & \xmark & \cmark  & 5.07 & 48.8 & 6.85 & 36.5\\
\textbf{UniScale} & \xmark & \cmark & 5.03 & 51.1 & 6.04 &  42.7\\
\hline

\multicolumn{7}{l}{\textbf{Multi-View w/ Alignment}} \\
\hline 
\multicolumn{7}{l}{\textbf{a) Images-only}} \\
No Prior Injection & \xmark & \xmark  & 3.51 & 64.7 & 1.69 & 86.7\\
No Prior Injection into Scale Head & \xmark & \xmark  & 3.48 & 65.5 & 1.69 & 86.7\\
No Scale Head & \xmark & \xmark  & 3.56 & 64.4  & 1.69 & 86.9\\
\textbf{UniScale} & \xmark & \xmark & 3.47 & 65.9 & 1.69 & 86.8 \\
\hline
\multicolumn{7}{l}{\textbf{b) Images+Poses+Intrinsics}} \\
No Prior Injection into Scale Head & \cmark & \cmark  & 3.75 & 61.9 & 1.73 & 85.9\\
No Scale Head & \cmark & \cmark  &  3.58 & 64.1 & 1.72 & 86.2 \\
\textbf{UniScale} & \cmark & \cmark & 3.50 & 65.3 & 1.71 & 86.2\\
\hline
\multicolumn{7}{l}{\textbf{c) Images+Intrinsics}} \\
No Prior Injection into Scale Head & \cmark & \xmark  & 3.59 & 64.6 & 1.73 & 86.1\\
No Scale Head & \cmark & \xmark  & 3.54 & 65.1 & 1.71 & 86.4\\
\textbf{UniScale} & \cmark & \xmark & 3.49 & 66.1 & 1.71 & 86.3 \\

\hline
\multicolumn{7}{l}{\textbf{d) Images+Poses}} \\
No Prior Injection into Scale Head & \xmark & \cmark  & 3.52 & 64.0 & 1.70 & 86.7\\
No Scale Head & \xmark & \cmark  &3.63 & 63.0 &1.70 & 86.7\\
\textbf{UniScale} & \xmark & \cmark & 3.51 & 64.7 &1.70 & 86.7 \\

\hline

\end{tabular}
\label{tab:ablation_prior}
\vspace{-4mm}
\end{table}

We investigate the impact of prior injection on overall model performance and evaluate our design choices for the prior encoders. Table~\ref{tab:ablation_prior} summarizes the quantitative results on the KITTI and ScanNet datasets.

\paragraph{Universal Model Training} We first evaluate the benefits of training UniScale as a universal model with probabilistic prior injection. Compared to a baseline trained without any prior injection, the universal model not only gains the flexibility to use priors when available but also yields improved performance in the image-only setting. This suggests that the model learns a more robust underlying geometric representation when trained to handle diverse input configurations.


\paragraph{Prior Injection in Scale Head} We further assess the impacts of injecting priors directly into the scale head. As shown in Table~\ref{tab:ablation_prior}, removing this information results in a notable performance drop across nearly all metrics. This confirms that conditioning the scale head on explicit prior cues is critical for accurate metric recovery, validating our split injection design.

\paragraph{Impact of Scale Head} We evaluate the contribution of the scale head itself by removing it entirely. Even under `Multi-View with Alignment' evaluation where absolute scale is factored out, we observe a drop in performance, indicating that the auxiliary task of explicit metric-scale prediction acts as a regularizer, benefits the learning of depth and point map estimations. 

\paragraph{Extrinsic Encodings} 

\begin{figure}[t]
  \centering
  \includegraphics[width=\linewidth]{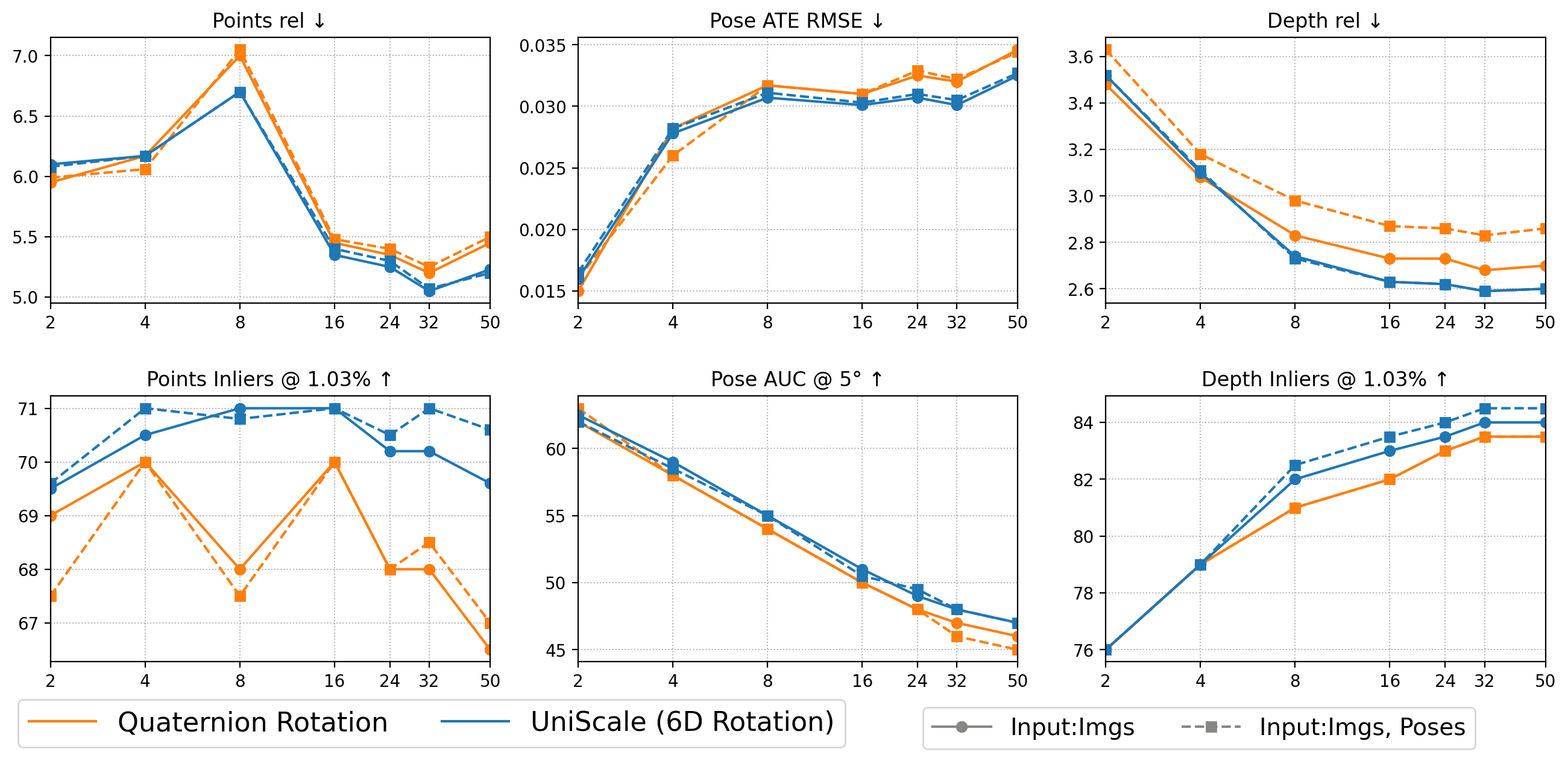}\vspace{-3mm}
  \caption{Comparison between 6D and Quaternion Encoding Strategies. 6D extrinsic encodings demonstrate better results on multi-view reconstruction for number of input views varying from 2 to 50 and under different input configurations.}\vspace{-5mm}
  \label{fig:quaternion_dense_n}
\end{figure}

To evaluate pose parameterization, we compare quaternion-based encoders with UniScale’s 6D rotation encoder on ETH3D~\cite{schops:etal:CVPR2017} and ScanNet++~\cite{dai2017scannet}. As shown in Fig.~\ref{fig:quaternion_dense_n}, both perform similarly with few views, but the 6D representation is more robust as the number of views increases, with a clear gap for $N \ge 8$.
We attribute this to the continuity of the 6D representation~\cite{zhou2019continuity}, which yields smoother optimization than discontinuous quaternions. In large-scale multi-view settings where rotation errors accumulate, this property is critical for stable global alignment, making 6D encoding preferable for scalable training.

\vspace{-1mm}
\section{Conclusion}
\vspace{0mm}
We present UniScale, a unified framework for metric 3D reconstruction suitable for robotic applications that recovers real-world scale and enables semantic-aware prior injection within a feed-forward architecture. A dedicated scale head overcomes normalization limits in existing models, while structured prior injection aligns geometric cues with their semantic roles. Extensive experiments show strong performance in depth estimation, calibration, and point cloud reconstruction, with seamless integration into unified frameworks. Its modular design allows UniScale to upgrade normalized reconstruction systems to metric ones. Future work includes extending UniScale to single-view settings and incorporating additional sensing modalities.


\bibliographystyle{IEEEtran}
\bibliography{main}

\end{document}